\documentclass{article}
\usepackage{ijcai18}

\usepackage{times}
\usepackage{soul}
\usepackage{url}
\usepackage[hidelinks]{hyperref}
\usepackage[utf8]{inputenc}
\usepackage[small]{caption}
\usepackage{float}
\usepackage{multirow}
\usepackage{amsmath}
\usepackage{amsthm}
\usepackage{array}
\usepackage{algorithm}
\usepackage[noend]{algpseudocode}
\usepackage{graphicx}
\usepackage{tikz}
\usepackage{pgfplots}
\usepackage{textcomp}
\usepackage{tikz}
\usetikzlibrary{shapes,backgrounds}

\graphicspath{ {images/} }

\usepackage{xcolor,colortbl}

\newcommand{\learner}{\textsc{Count-OR}}
\newcommand{\TX}{\textbf{X}}
\newcommand{\TY}{\textbf{Y}}
\newcommand{\TZ}{\textbf{Z}}
\newcommand{\Nonzero}[2]{\text{Nonzero}\!\left(#1,#2\right)}
\newcommand{\Sum}[2]{\text{Sum}\!\left(#1,#2\right)}

\newcommand{\Count}[3]{\text{Count}\!\left(#1,#2,#3\right)}

\newcommand{\Nurses}{\texttt{Nurses}}
\newcommand{\Shifts}{\texttt{Shifts}}
\newcommand{\Days}{\texttt{Days}}
\newcommand{\Nurse}{\texttt{Nurse}}
\newcommand{\Shift}{\texttt{Shift}}
\newcommand{\Day}{\texttt{Day}}

\usepackage{etoolbox}
\let\bbordermatrix\bordermatrix
\patchcmd{\bbordermatrix}{8.75}{4.75}{}{}
\patchcmd{\bbordermatrix}{\left(}{\left[}{}{}
\patchcmd{\bbordermatrix}{\right)}{\right]}{}{}

\definecolor{Colored}{rgb}{0.85,0.85,0}
\newcolumntype{a}{>{\columncolor{Colored}}c}

\title{Automating Personnel Rostering by Learning Constraints Using Tensors}
\author{
Mohit Kumar$^1$,
Stefano Teso$^1$,
Luc De Raedt$^1$
\\ 
$^1$ KU Leuven\\
\texttt{\{mohit.kumar,stefano.teso,luc.deraedt\}@cs.kuleuven.be}
}

\begin{document}

\maketitle

\begin{abstract}
Many problems in operations research require that constraints be specified in the model. Determining the right constraints is a hard and laborsome task.
We propose an approach to automate this process using artificial intelligence and machine learning principles. So far there has been only little work on learning constraints within the operations research community.
We focus on personnel rostering and scheduling problems in which there are often past schedules available and show that it is possible to automatically learn  constraints from such examples. 
To realize this, we adapted some techniques from the constraint programming community and  we have extended them in order to  cope with multidimensional examples. 
The method uses a tensor representation of the example, which helps in capturing the dimensionality as well as the structure of the example, and applies tensor operations to find the constraints that are satisfied by the example. 
To evaluate the proposed algorithm, we used constraints from the Nurse Rostering Competition and generated solutions that satisfy these constraints; these  solutions were then used as examples to learn constraints. Experiments demonstrate that the proposed algorithm is capable of producing human readable constraints that capture the underlying characteristics of the examples. 
\end{abstract}

\section{Introduction}
\label{sec:intro}

Constraints are pervasive in practical scheduling and rostering problems. For example, hospitals usually generate a weekly schedule for their nurses based on constraints like the maximum number of working days for a nurse. As the number of nurses and the complexity of the constraints increases, however, generating the schedule manually becomes impossible.  Organizations may hire domain experts to manually model the constraints, but this is expensive and time consuming. A tempting alternative is to employ constraint learning~\cite{de2018learning} to automatically induce the constraints from examples of past schedules.

Unfortunately, existing constraint learners are not tailored for this setting.  Classical approaches like Conacq~\cite{Bessiere} and Inductive Logic Programming tools~\cite{MUGGLETON1994629} focus on logical variables only,  while scheduling constraints often include numerical terms.  Very few approaches can handle this case.  TaCLe~\cite{kolb2017learning} focuses on 2-D tabular data (Excel spreadsheets), while schedules are inherently multi-dimensional. To see what we mean, consider the nurse schedule shown in Table~\ref{tab:dataformat}.  For each combination of nurse, day and shift the value of 1 represents that the nurse worked in that particular shift of that day, while a 0 means the nurse didn't work.  It is easy to see that nurses, days, and shifts are independent of each and behave like different dimensions.  ModelSeeker~\cite{beldiceanu2012model} is the only method that can handle such multi-dimensional structures, but it is restricted to global constraints only.


To address this issue, we propose COnstraint UsiNg TensORs (\learner{}), a novel constraint learning approach that leverages tensors for capturing the inherent structure and dimensionality of the schedules.  In order to learn the constraints, \learner{} extracts and enumerates all (meaningful) slices of the input schedule(s), aggregates them through tensor operations, and then computes bounds for the aggregates to generate candidate numerical constraints.  Some simple filtering strategies are applied to prune irrelevant and trivially satisfied candidates.  When increasing the number of dimensions in the example, the rank of the tensor representing the example will increase accordingly but the proposed method \learner{} remains unchanged, so \learner{} can easily scale to large and complex schedules. The number of candidate sub-tensors however increases exponentially with the number of dimensions of $\TX$.

We make the following key contributions:
(\textbf{1}) A tensor representation of schedules and constraints appropriate for real-world personnel rostering problems.
(\textbf{2}) A novel constraint learning algorithm, \learner{}, which uses tensor extraction and aggregation operations to learn the constraints hidden in the input schedules.
%
(\textbf{3}) An empirical evaluation on real-world nurse rostering problems.
%

The paper is structured as follows. We present the method in Section~\ref{sec:method}, followed by evaluation on example instances  in Section~\ref{sec:experiments}. We conclude with some final remarks in Section~\ref{sec:conclusion}.

\section{Method}
\label{sec:method}

At a high level, \learner{} consists of four steps.  Given one or more input schedules:
(1) In a first step, each input schedule is mapped to a tensor whose elements represent, in the simplest case, which employees worked in which time slot.
(2) Next, the algorithm enumerates all sub-tensors and summarizes them by applying one or more aggregation operations.  This way it obtains several quantities of interest, for instance the number of employees each day or the number of working days for each employee.
(3) Numerical bounds for these quantities are then computed by taking the minimum and maximum across all sub-tensors, producing a number of candidate constraints of the form ``the minimum number of employees each day is 4''.
(4) Finally, trivially satisfied candidates are filtered out to produce a set of consistent constraints.  These can be readily fed to any constraint solver to generate new schedules consistent with the input examples.
We proceed by describing these steps in detail.

\textbf{(1) Mapping schedules to tensors.} Consider the example nurse schedule in Table~\ref{tab:dataformat}.  Here the rows represent different nurses and the columns represent shifts on different days.  A value of 1 means that the corresponding nurse worked in that shift on that particular day, while 0 means that the nurse did not work.  This is a very natural representation for schedules.
More formally, let $D = \{D_1, D_2, \ldots, D_n\}$ be the scheduling \emph{dimensions} and $D_i$ the set of distinct values for $i$th dimension.  In our example schedule there are 3 dimensions, namely $D = \{\Nurses, \Days, \Shifts\}$, and $\Nurses = \{\Nurse_1, \ldots, \Nurse_4\}$.
%
A schedule in this format can be readily represented by a rank-$n$ \emph{tensor}, where $n = |D|$ is the number of dimensions in the schedule.  The shape of the tensor reflects the number of distinct values for each dimension.  In our example, the schedule has 3 dimensions, so it can be represented by a tensor of rank $3$ with shape $[4, 7, 3]$.  The elements of the tensor are identified by a particular (nurse, day, shift) combination.  For instance, $\TX[\Nurse_2, \Day_1, \Shift_3]$ shows whether $\Nurse_2$ worked on $\Day_1$ in $\Shift_3$.

\begin{table}[t]
\begin{small}
    \begin{center} 
        \begin{tabular}{c|a|a|a|c|c|c|c|c|c|}
            \cline{2-10}
            \multicolumn{1}{l|}{}                  & \multicolumn{3}{c|}{$\text{Day}_1$} & \multicolumn{3}{c|}{$\text{Day}_2$} & \multicolumn{3}{c|}{$\text{Day}_3$} \\
            \cline{2-10}
            \multicolumn{1}{l|}{}                  & $\text{S}_1$     & $\text{S}_2$     & $\text{S}_3$     & $\text{S}_1$     & $\text{S}_2$     & $\text{S}_3$     & $\text{S}_1$     & $\text{S}_2$     & $\text{S}_3$ \\
            \hline
            \multicolumn{1}{|c|}{\cellcolor{Colored}$\text{Nurse}_1$} &  1      &  0      &  0      & 0      & 1      & 0      & 0      & 1      & 0 \\
            \hline
            \multicolumn{1}{|c|}{\cellcolor{Colored}$\text{Nurse}_2$} & 0      & 1      & 0      & 1      & 0      & 0      & 0      & 0      & 1 \\
            \hline
            \multicolumn{1}{|c|}{\cellcolor{Colored}$\text{Nurse}_3$} & 0      & 0      & 1      & 0      & 0      & 0      & 1      & 0      & 0 \\
            \hline
            \multicolumn{1}{|c|}{\cellcolor{Colored}$\text{Nurse}_4$} & 0      & 0      & 1      & 0      & 0      & 1      & 1      & 0      & 0 \\
            \hline
        \end{tabular}
    \end{center}
    \caption{\label{tab:dataformat} Example nurse schedule with three dimensions: four nurses, three days, and three shifts per day.}
\end{small}
\end{table}

Before proceeding, we introduce some required notation.  We write $\TX[d_i]$ to indicate the sub-tensor obtained by fixing the $i$th dimension of $\TX$ to $d_i \in D_i$. For example, $\TX[\texttt{Day}_1]$ extracts the working schedule of all nurses for the three shifts (shown in yellow in Table~\ref{tab:dataformat}).  Fixing multiple dimensions is also allowed.  For instance, $\TX[\Nurse_1, \Day_1]$ extracts the sub-tensor $[1, 0, 0]$, where the elements refer to the different shifts for $\Nurse_1$ in $\Day_1$.  We will make extensive use of the Cartesian product, defined as
$D_1 \otimes D_2 = \{(d_1, d_2) \mid d_1 \in D_1, d_2 \in D_2 \}$.
For any choice of dimensions $D' \subseteq D$, we use the shorthand $\bigotimes(D')$ to indicate the Cartesian product of the dimensions in $D'$.
%
%

\textbf{(2) Enumeration and aggregation.} We are now ready to introduce the first two aggregation functions, Nonzero and Sum, which play a central role in our algorithm.  Given an input tensor $\TX$, $\Nonzero{\TX}{D'}$ reduces it to an aggregate tensor $\TY$ indexed by $e \in \bigotimes(D')$, by checking for each $e$ whether there exists at least one non-zero element in the sub-tensor $\TX[e]$.  More formally, letting $I(c)$ be the indicator function,
the output of Nonzero satisfies $\TY[e] = I(\TX[e] \ne \textbf{0})$ for every $e \in \bigotimes(D')$.  The $\text{Sum}(\TX, D')$ function is defined analogously, except that $\TY$ is obtained by summing up all the elements of $\TX[e]$, that is, $\TY[e] = \text{sum of values in} \;\TX[e]$ for all $e$.
These functions allow us to capture many quantities of interest.  For instance, checking whether $\Nurse_i$ works in \emph{any} shift of $\Day_j$ can be accomplished by applying the Nonzero function with $D' = \{\Nurses, \Days\}$, so that $\TY$ is:
$$ \TY[\Nurse_i, \Day_j] = I(\TX[\Nurse_i, \Day_j] \ne \textbf{0}) $$
Similarly, the number of shifts worked by $\Nurse_i$ on $\Day_j$ can be retrieved by applying the Sum function with $D' = \{\Nurses, \Days\}$, so that $\TY$ is:
$$ \TY[\Nurse_i, \Day_j] = \text{sum of values in $\TX[\Nurse_i, \Day_j]$} $$
By varying $D'$, these functions produce other relevant quantities, such as the number of working employees for each day, the number of working days for each employee, \emph{etc}.

We introduce one more function, Count, which combines Nonzero and Sum to express even more quantities of interest.  Let $M$ and $S$ be two disjoint, non-empty subsets of $D$.  Count is defined as:
\[
    \label{eq:count}
    \Count{\TX}{M}{S} = \Sum{\Nonzero{\TX}{M\cup S}}{S}
\]
For instance, for $M = \{\Nurses\}$ and $S = \{\Days\}$, $\Nonzero{\TX}{M \cup S}$ returns a 2-d tensor $\TY$, of shape [4, 3] over the 4 nurses and the 3 days, where $\TY[\Nurse_i,\Day_j]$ encodes whether the $i$th nurse worked in any shift on the $j$th day.  Count then applies Sum to this tensor to obtain a 1-d tensor, $\TZ = \Sum{\TY}{S}$, of shape [3] over the three days, where $\TZ[\Day_i]$ encodes the total number of nurses working on the $i$th day.  The end result is that the tensor output by $\Count{\TX}{M}{S}$ encodes the total number of distinct employees working on different days: $\TZ = [4 \;\; 3 \;\; 4]$.

\textbf{Dealing with other constraints.}  By definition, the Sum and Nonzero ignore the order in which zeros and ones appear in the input tensor.  Order, however, is essential for capturing quantities like ``the minimum (or maximum) number of consecutive holidays for an employee'' or ``the maximum number of consecutive working days (or shifts) for an employee''.  To deal with these, we introduce four more aggregation functions: MinConsZero, MinConsOne, MaxConsZero, and MaxConsOne, described next.

Given an input tensor $\TX$ and a subset of dimensions $D'$, $\text{MaxConsOne}(\TX, D')$ outputs a tensor $\TY$ of rank $|D'|$ by taking maximum number of consecutive ones in $\TX[e]$, where $e \in \bigotimes(D')$.  Similarly, MinConsOne computes the minimum number of consecutive ones.  These two functions produce an upper and lower bound, respectively.  The two other functions, MaxConsZero and MinConsZero work analogously, but for consecutive zeros.

To see how these work, consider the case $M = \{\Days\}$ and $S = \{\Nurses\}$: replacing Sum with MaxConsOne in Eq.~\ref{eq:count} allows us to compute number of maximum consecutive working days for each nurse.  Taking the maximum of these values give us an upper bound, the maximum consecutive working days for any nurse.

Note that when $\TX[e]$ is a tensor of rank $\ge 1$ (i.e., when it is not a vector), talking about consecutive ones doesn't make much sense.  So for these four functions we only consider the cases where $\TX[e]$ is a 1-d tensor, i.e., $|D \setminus D'| = 1$.

\textbf{(3) Computing the bounds.}  After enumerating all the quantities of interest, \learner{} computes their minima and maxima to produce candidate constraints capturing their variation. So for each combination of $M$ and $S$, first it calculates $\Count{\TX}{M}{S}$, from there it computes the bounds as $Minimum(\Count{\TX}{M}{S}) \le \Count{\TX}{M}{S} \le Maximum(\Count{\TX}{M}{S})$.
For instance, when applied to our example schedule, our algorithm would produce the lower and upper bounds for each candidate constraints listed in Table~\ref{tab:constraints} (among others).

\begin{table}[tb]
    \begin{center} \begin{small}
        \begin{tabular}{|m{7em}|m{7em}|m{7em}|}
            \hline
            $M$                 & $S$                 & $\Count{\TX}{M}{S}$                                     \\ \hline
            \{\Days\}        & \{\Nurses\}          & \# of working days / Nurse                                         \\ \hline
            \{\Days, \Shifts\}        & \{\Nurses\}   & \# of working shifts / Nurse                                        \\ \hline
            \{\Nurses\}          & \{\Days\}        & \# of distinct employees / day                              \\ \hline
            \{\Shifts\}          & \{\Days\} & \# of shifts for each day with at least one nurse working                                    \\ \hline
            \{\Shifts\}   & \{\Nurses, \Days\}        & \# of working shifts per day per nurse                               \\ \hline
            \{\Days\}  & \{\Nurses, \Shifts\}         & \# of working days in the same shift / nurse                     \\ \hline
            \{\Nurses\}    &   \{\Days, \Shifts\}     & \# of nurses / shift each day                                 \\ \hline
        \end{tabular}
    \end{small} \end{center}
    \caption{\label{tab:constraints} A selection of scheduling constraints representable by the Count function for different choices of $M$ and $S$.}
\end{table}

\textbf{(4) Dealing with irrelevant constraints.}  To learn the candidate constraints, we consider all possible combinations of $M$ and $S$ and then apply aggregation functions to the input tensor.  Therefore, no matter what the input is, we will always obtain candidates for each constraint represented by our mathematical equations. 
In this process we might acquire some trivially satisfied or meaningless constraints.  
For example, learning that the minimum number of working days for a nurse is 0 every week doesn't give any information at all, because the number of working days is always non-negative hence the learned constraint is trivially satisfied. 

We might also learn some meaningless constraints. Depending on the application domain, some combinations of $M$ and $S$ may lead to formally sound, but meaningless, constraints. For example, if $M=\{\Shifts\}$ and $S=\{\Days\}$, the count we get represents ``the number of shifts for each day where there was at least one nurse working'', which doesn't make much sense. 
The fact that such combinations should be avoided can be introduced as background knowledge into our algorithm.  When instructed as such, \learner{} does not enumerate sub-tensors for the known meaningless choices of $M$ and $S$.  This can be partially automated by observing some common patterns followed by the meaningless constraints. 
Specifically, for constraints learned using the advanced functions, we know that in our examples ordering of Nurses is not important as all the nurses are identical, so finding consecutive values of ones or zeros across Nurses doesn't make any sense.  So we use the following method 
to neglect such constraints.  For a constraint, if $D_i \in M$ such that ordering of values in $D_i$ is not important, then we neglect the constraint learned using advanced functions.

Second, we filter out trivially satisfied candidate constraints as follows.  For the upper bound, if it holds that:
$$
	\textstyle
    Maximum(\Count{\TX}{M}{S}) = |\bigotimes(S)|
$$
then we discard this upper bound because $|\bigotimes(S)|$ is the maximum possible value $\Count{\TX}{M}{S}$ can take so this bound is implicit.  
For example, when $M=\{\Nurses\}$ and $S=\{\Days, \Shifts\}$, if we learn that $Maximum(\Count{\TX}{M}{S}) = |\bigotimes(S)| = 21$, it translates to maximum number of working shifts being 21 in a week using Eq.~\ref{eq:count}, which is an obvious bound, so we will drop this constraint.  Second, for the lower bound, if
$$
    Minimum(\Count{\TX}{M}{S}) = 0
$$
then again we drop this constraint as it's an obvious lower bound.
These same two rules hold when using advanced functions instead of $Sum$ in Equation~\ref{eq:count}.

These simple filtering rules can be surprisingly effective in practice.  In our experiments with real-world scheduling problems, we noticed that they get rid of most of the irrelevant constraints, without having to tell explicitly the algorithm which constraints to ignore.  This is because for many such constraints the bounds learned either have the maximum possible or minimum possible value. So they get filtered when the above two filtering rules are applied. 


\section{Empirical Analysis}
\label{sec:experiments}

\begin{figure*}[tb]
    \centering
    \begin{tabular}{cc}
        \includegraphics[width=0.3\textwidth]{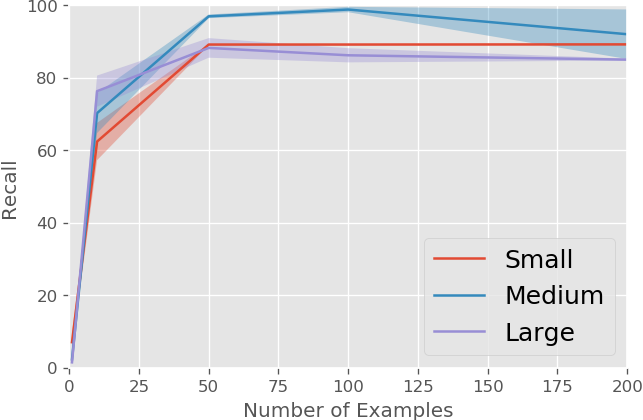}
        &
        \includegraphics[width=0.3\textwidth]{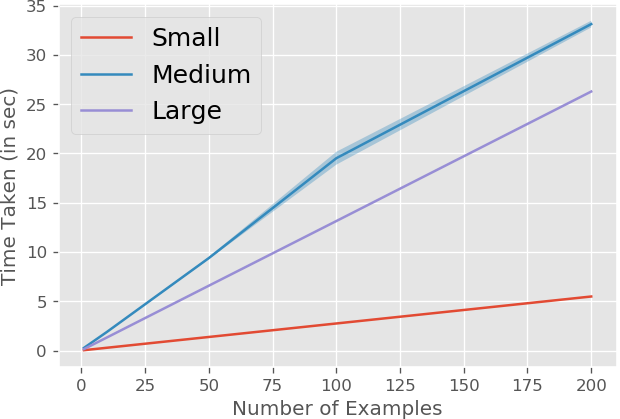}
    \end{tabular}
    \caption{\label{fig:results} Left: recall vs number of examples used. Right: time taken to learn the constraints vs number of examples used.}
\end{figure*}

In this section we empirically evaluate whether \learner{} can recover a target scheduling model taken from the Second International Nurse Rostering Competition\footnote{URL: \texttt{http://mobiz.vives.be/inrc2/}} (INRC-II~\cite{sara2014second}).  In particular, we address the following research questions:
(\textbf{Q1}) Do the learned constraints produce schedules similar to those produced by the target model?
(\textbf{Q2}) How many example schedules does the learner need to achieve this?


We chose a target model $M_T$ that only includes constraints that can be exactly represented in our language, for simplicity.  The model includes 11 such constraints.  We used the Gurobi solver to generate 10,000 solutions to $M_T$, and used increasingly larger subsets of this sample as input to \learner{}.  Then we evaluated the model learned by the algorithm, $M_L$, by measuring its precision and recall with respect to the target model $M_T$:
$$
    \text{pr} = \frac{|\text{Sol}(M_T) \cap \text{Sol}(M_L)|}{|\text{Sol}(M_L)|}
    \;\;
    \text{rc} = \frac{|\text{Sol}(M_T) \cap \text{Sol}(M_L)|}{|\text{Sol}(M_T)|}
$$
Here $\text{Sol}(M) = \{ x \,:\, M \models x \}$ is simply the set of solutions of model $M$.  Computing these quantities is not trivial, so we estimate them using sampling.  To compute the recall, we generate 10,000 examples, randomly choose different number of examples from this set and learn the constraints using \learner{} and then check how many of these 10,000 examples satisfy the learned model.  The precision is computed analogously.  

We evaluated \learner{} on three different scenarios by changing the number of nurses and bounds for some constraints.  This setup is meant to simulate hospitals of different sizes: a small hospital (10 nurses, 28 days, 4 shifts), a medium sized hospital (31, 28, 4), and a large hospital (49, 28, 4). To design these models, we used ``sprint'', ``medium'' and ``large'' example instances from Nurse rostering competition\footnote{URL: \texttt{https://bit.ly/2IyObjD}}, which represent models of realistic sizes. 
To address \textbf{Q2}, the number of examples used was also varied.

Recall that, in this setting, all the constraints in the target model are representable by our constraint language.  As a consequence, \learner{} always achieves 100\% precision for all target models.  The recall, reported in Figure~\ref{fig:results} (left), is more interesting.  The $x$-axis is the number of input examples provided to \learner{}, while the $y$-axis is the average recall computed over 5 different subsets of examples. We can see that recall is not very good when learning from just 1 example, but as we increase the number of examples we achieve $\sim90\%$ recall in all the cases. The time taken to learn the constraints, see Figure~\ref{fig:results} (right), increases linearly with the number of examples, and for 50 examples, when the recall is $\approx 90\%$, the time taken on average is around 6 seconds.



\section{Conclusion}
\label{sec:conclusion}

We proposed a novel constraint learning approach, named \learner{}, specifically tailored for personnel rostering problems.  Given examples of past schedules, \learner{} leverages tensors and operations on them to uncover the constraints hidden in the examples.  The key idea is that many quantities of interest, like maximum working hours or the number of employees needed each day, can be readily computed by applying aggregation operators to a tensor representation of the example schedules.  This naturally takes into account the intrinsic structure and dimensionality of the scheduling problem.  Our empirical evaluation shows that \learner{} can easily and quickly recover the constraints appearing in real-world nurse rostering problems.

We are working on two generalizations of \learner{}.  First, we plan to introduce support for background knowledge.  Potential applications include encoding the skills of the employees and the skills required for performing certain tasks; the learned constraints then would automatically exploit, and be consistent with, this additional knowledge.  We plan to use tensor products to introduce this information into the problem.  Second, we are currently working on extending \learner{} to other OR applications, for instance scheduling sport matches and tournaments.  This involves introducing new application-specific aggregation operators.

\paragraph{Acknowledgements.} This work has received funding from the European Research Council (ERC) under the European Union’s Horizon 2020 research and innovation programme (grant agreement No [694980] SYNTH: Synthesising Inductive Data Models)”.

\bibliographystyle{unsrt}
\bibliography{ijcai18}
\end{document}